\documentclass{article}

\usepackage{microtype}
\usepackage{multirow}
\usepackage{graphicx}
\usepackage{subfigure}
\usepackage{booktabs} 

\usepackage{hyperref}

\usepackage[accepted]{icml2021}
\usepackage[table,xcdraw]{xcolor}
\usepackage{xr}

\icmltitlerunning{Predicting Mood Disorder Symptoms using Interpretable Multimodal Dynamic Attention Fusion Network}

\newcommand{\beginsupplement}{%
        \setcounter{table}{1}
        \renewcommand{\thetable}{S\arabic{table}}%
        \setcounter{figure}{1}
        \renewcommand{\thefigure}{S\arabic{figure}}%
        \setcounter{section}{1}
        \renewcommand{\thesection}{S\arabic{figure}}%
     }

\begin{document}

\twocolumn[
\icmltitle{Predicting Mood Disorder Symptoms with Remotely Collected Videos Using an Interpretable Multimodal Dynamic Attention Fusion Network}



\icmlsetsymbol{equal}{*}

\vspace{-0.3cm}
\begin{icmlauthorlist}
\icmlauthor{Tathagata Banerjee}{equal,to}
\icmlauthor{Matthew Kollada}{equal,to}
\icmlauthor{Pablo Gersberg}{to}
\icmlauthor{Oscar Rodriguez}{to}
\icmlauthor{Jane Tiller}{to}
\icmlauthor{Andrew E Jaffe}{to}
\icmlauthor{John Reynders}{to}
\end{icmlauthorlist}

\centering$^*$Equal contribution, $^1$BlackThorn Therapeutics, San Francisco,California, USA. \\ 
\centering Email: tathagata.banerjee@blackthornrx.com

\vspace{-0.1cm}



\icmlkeywords{Multimodal Machine Learning, Mood Disorders, Remote Data, }

\vskip 0.3in
]



\printAffiliationsAndNotice{} 

\begin{abstract}
We developed a novel, interpretable multimodal classification method to identify symptoms of mood disorders viz. depression, anxiety and anhedonia using audio, video and text collected from a smartphone application. We used CNN-based unimodal encoders to learn dynamic embeddings for each modality and then combined these through a transformer encoder. We applied these methods to a novel dataset - collected by a smartphone application - on 3002 participants across up to three recording sessions. Our method demonstrated better multimodal classification performance compared to existing methods that employed static embeddings. Lastly, we used SHapley Additive exPlanations (SHAP) to prioritize important features in our model that could serve as potential digital markers.
\end{abstract}


\section{Introduction} \label{intro}
Mood disorders are common and debilitating, and can present across multiple related symptom axes, including depressive, anxious, and anhedonia symptoms. Developing scalable tools that can help researchers better quantify symptoms of these complex disorders is a promising research avenue towards improved characterization, diagnosis, and subsequent care.
Since mood disorder symptoms may manifest in patient speech and facial expressions \cite{cummins2015,Pampouchidou2019}, this research can benefit from the progress that deep learning has heralded in modeling audio, video and text data, for example using remotely collected videos.
Improvements in symptom characterization requires machine learning methods that are (1) \emph{multimodal}: successful in integrating information from different modalities (2) \emph{interpretable}: can be validated by clinicians and scientists
(3) \emph{robust}: can handle data that is collected through methods that scale (e.g. remote collection) and (4) \emph{transdiagnostic}: successful in identifying symptoms in different mood disorders. 
Despite the substantial progress made in this field, including through the AVEC workshops \cite{Ringeval2019,Ringeval2017,Valstar2016}, as well as other related influential publications \cite{AlHanai2018,haque2018}, there are nevertheless  several existing gaps: (1) The multimodal fusion methods developed for this application have relied on fusing static embeddings, which do not integrate temporal information across modalities, potentially inhibiting performance. (2) There has not been much focus on model interpretability such as feature importance which can help clinicians validate such methods and provide pathways for digital marker discovery. (3) The data used in the existing literature has been collected in clinical or laboratory settings, thus limiting adoption to more scalable, but potentially noisier data collection methods. (4) The existing literature has focused only on symptoms of depression, but mood disorders may consist of other co-morbid illnesses like anxiety and/or anhedonia.

\textbf{Contributions} 
We propose a novel, interpretable, multimodal machine learning method to identify symptoms of mood disorders using audio, video and text collected using a smartphone app. We addressed the aforementioned issues by (1) developing a multimodal method that took into account dependencies across modalities within a temporal context by learning dynamic (i.e. temporal) embeddings for each modality and then combining them through a transformer-based multimodal fusion model. This framework showed improved performance compared to previous approaches that employ static (i.e. non-temporal) unimodal embeddings and/or use fewer modalities. (2) We used SHapley Additive exPlanations (SHAP) to identify the most important features in these models. (3) We evaluated these methods using a novel dataset that we collected remotely through consumer smartphones without clinical supervision (after developing robust quality control methods to reduce potential variability). (4) Our dataset included questionnaires for multiple mood disorder symptoms including depression, anxiety, and anhedonia, which we predicted using our multimodal modeling.

\section{Data} \label{data}
The data used in this paper was collected remotely through an interactive smartphone application 
that was available to the U.S. general population through Google Play and Apple App Store under IRB approval. 
(1) Demographic Variables and Health History (2) Self-reported clinical scales including Patient Health Questionnaire-9 (PHQ-9) \cite{Kroenke2002}, Generalized Anxiety Disorder-7 (GAD-7) \cite{Spitzer2006} and Snaith Hamilton Pleasure Scale (SHAPS) \cite{Snaith1995} and (3) Video recorded vocal expression activities (where participants were asked to record videos of their faces while responding verbally to prompts) were collected on each of 3002 unique participants. The entire set of video tasks took less than five minutes, and participants could provide data up to three times (across 4 weeks), for a total of 3 sessions (not all participants completed 3 sessions).  


\section{Feature Extraction and Quality Control} \label{feqc}
Audio, video and text features were extracted to perform model building. However, since this data was collected without human supervision, a rigorous quality control procedure was performed to reduce noise. Additional details are provided in the Supplement.
\subsection{Feature Extraction}

\textbf{Audio:} These represent the acoustic information in the response. Each audio file was denoised, and unvoiced segments were removed. A total of 123 audio features (including prosodic, glottal and spectral) \cite{cummins2015} were extracted at a resolution of 0.1 seconds. 


\textbf{Video:} These represent the facial expression information in the response. For each video, 3D facial landmarks were computed at a resolution of 0.1 seconds. From these, 22 Facial Action Units \cite{ekman2002} were computed for modeling.

\textbf{Text:} These represent the linguistic information in the response. Each audio file was transcribed using Google Speech-to-Text and 52 text features were computed including affective features, word polarity and word embeddings  \cite{Warriner13,blob,le14}.

\subsection{Quality Control} 
In contrast to prior literature, where the data was collected under clinical supervision (e.g. the DAIC-WOZ dataset \cite{gratch-etal-2014-distress}), our data was collected remotely on consumer smartphones. Consequently, this data could have more noise that needed to be addressed before modeling. There were two broad sources of noise: (1) Noisy medium (e.g. background audio noise, video failures and illegible speech) (2) Insincere participants (e.g. participant answering ``blah'' to all prompts). 
Using the metadata, scales and extracted features, we designed quality control flags to screen participants.  Out of 6020 collected sessions, 1999 passed this stage. The developed flags can be pre-built into the app for data collection in future.


\section{Machine Learning Problem}  \label{mlp}

We took a multimodal machine learning approach to classify symptoms of mood disorders. Specifically, we used the audio, video and textual modalities for the 1999 sessions as input, and performed three classification problems to predict binary outcome labels related to the presence of symptoms of (1) depression (total PHQ-9 score $>$ 9), (2) anxiety (total GAD-7 score $>$ 9), and (3) anhedonia (total SHAPS score $>$ 25)). Here, 71.4\% of participants had symptoms of depression, 57.8\% of participants had symptoms of anxiety and 67.3\% of participants had symptoms of anhedonia. Our dataset was much larger than the DAIC-WOZ dataset in AVEC 2019 (N=275) \cite{Ringeval2019} and also contained a higher percentage of individuals with depression symptoms (our dataset=71.4\%, AVEC=25\%).


\begin{figure*}[ht] 
\centering
    \includegraphics[width=1.0\textwidth]{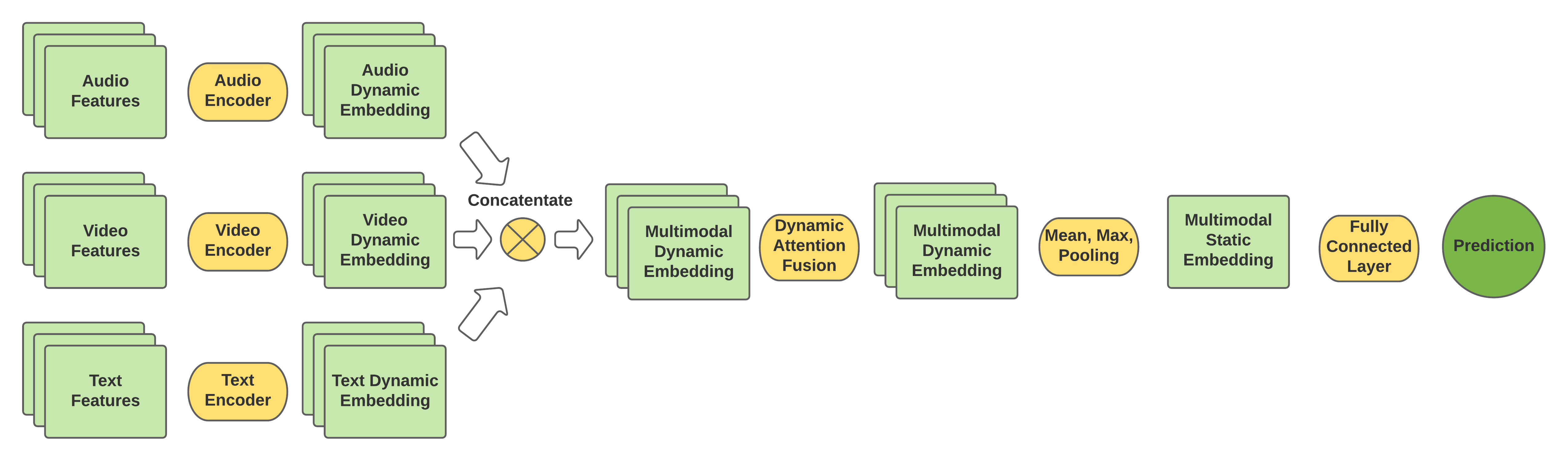}
    \caption{Diagram of Multimodal Classification System Architecture.}
    \label{figmulti}
\end{figure*}

\section{Model} \label{model}

\subsection{Model Architecture}
We proposed a novel multimodal classification framework: Unimodal encoders of stacked convolution blocks (1D convolution layer followed by batch normalization and ReLU activation) extracted dynamic embeddings from each of the individual modalities. These were concatenated and passed on to a Transformer Encoder \cite{vaswani17} to learn a multimodal representation for classification. Our architecture was in contrast to existing literature which used LSTM-based unimodal embeddings instead of convolution, which are then collapsed across the time dimension prior to fusion, as opposed to our dynamic fusion \cite{AlHanai2018,haque2018,ray2019,Qureshi2019}. This allowed us to integrate information across modalities within a temporal context (e.g. facial expression change and vocal inflection while a salient word is uttered) through dynamic embeddings to form a multimodal sequence and efficiently process this sequence using a transformer. The complete architecture is shown in Figure \ref{figmulti}. Additional details are provided in the Supplement.

\subsection{Model Training and Evaluation}
The unimodal encoders were pre-trained on the labels; the dynamic embeddings from these were fused using the Transformer Encoder and used for model training. We used a 60-20-20 random train-validation-test split and computed the F1 score for each split. This iteration was repeated 100 times and we reported the median test F1 score as the evaluation metric. Additional details are provided in the Supplement.

\subsection{SHAP-based Feature Importance}

One of the key issues in machine learning systems for healthcare is lack of interpretability, particularly in understanding feature importance. This is particularly the case for deep models like those we use in this study, which are useful in extracting information from rich modalities like audio, video, and text, but are difficult to interpret. Improvements in interpretability would open up the models to validation by clinicians, and might provide a pathway for digital marker discovery. We used SHapley Additive exPlanations (SHAP) \cite{lundberg2017}, an additive feature attribution method to explain the output of any machine learning model (not to be confused with the SHAPS questionnaire for assessing anhedonia). This enabled us to compute feature importance for each input variable using SHAP values.


\section{Experiments and Results} \label{result}

\subsection{Multimodal Classification of Symptom Severity}

Our \textbf{CNN-Dynamic Attention Fusion multimodal} method outperformed state of the art works employing static unimodal embeddings: (A) \textbf{LSTM-Concatenation} \cite{AlHanai2018} (B) \textbf{BiLSTM-Static Attention} \cite{ray2019} (C) \textbf{LSTM-Tensor Fusion} \cite{Qureshi2019} in multimodal classification of symptoms across three domains: depression (PHQ-9), anxiety (using GAD-7) and anhedonia (using SHAPS). Two different aspects of performance were compared. 

(1) We compared the overall classification performance across the three scales (using median test F1 score as the metric) in Table \ref{table1}. Our method performed better compared to the other methods across the three scales; it outperformed the next best model (i.e. LSTM-Tensor Fusion) by 4.94 \%, 3.60 \% and 3.98 \% for PHQ-9, GAD-7 and SHAPS respectively.

(2) We built models with each of the modalities and compared the performance of the multimodal model vs the best unimodal model (using the percentage difference in median test F1 score between multimodal and best unimodal) for the different approaches and across the three scales (Table \ref{table2}). Our method showed a notable increase in performance in the multimodal case whereas the other approaches showed very minor increase (or sometimes decrease). This demonstrated that our method was able to efficiently capture the supplementary information across different modalities. Complete results for all models are shown in the Supplement.

\begin{table}[t] 
\centering
\caption{Multimodal classification of mood disorder symptoms: Median Test F1 Score}
\label{table1}
\vskip 0.15in
\begin{tabular}{|c|c|c|c|}
\hline
\textbf{\begin{tabular}[c]{@{}c@{}}Fusion Approach 
\end{tabular}}            & \textbf{PHQ-9} & \textbf{GAD-7} & \textbf{SHAPS} \\ \hline
\begin{tabular}[c]{@{}c@{}}LSTM-Concatenation \end{tabular}                & 0.638          & 0.598          & 0.578          \\ \hline
\begin{tabular}[c]{@{}c@{}}BiLSTM-Static Attention \end{tabular}             & 0.625          & 0.5716         & 0.566          \\ \hline
\begin{tabular}[c]{@{}c@{}}LSTM-Tensor Fusion \end{tabular} 
                                                             & 0.632          & 0.601         & 0.567         \\ \hline
\begin{tabular}[c]{@{}c@{}}\textbf {CNN-Dynamic Attention} \end{tabular} & \textbf{0.664} & \textbf{0.623} & \textbf{0.590} \\ \hline
\end{tabular}
\vskip -0.1in
\end{table}

\begin{table}[] 
\centering
\caption{Percentage Difference in Median Test F1 Score between trimodal and best unimodal model}
\label{table2}
\vskip 0.15in
\begin{tabular}{|c|c|c|}
\hline
\textbf{Scale}                  & \textbf{Approach}                                                        & \textbf{\begin{tabular}[c]{@{}c@{}}Percentage \\ Difference\end{tabular}} \\ \hline
\multirow{4}{*}{\textbf{PHQ-9}} & LSTM-Concatenation                                                       & -0.79                                                                     \\ \cline{2-3} 
                                & \begin{tabular}[c]{@{}c@{}}BiLSTM-Static Attention\end{tabular}        & 0                                                                         \\ \cline{2-3} 
                                & \begin{tabular}[c]{@{}c@{}}LSTM-Tensor Fusion\end{tabular}             & 0.16                                                                      \\ \cline{2-3} 
                                & \textbf{\begin{tabular}[c]{@{}c@{}}CNN-Dynamic Attention\end{tabular}} & \textbf{3.05}                                                             \\ \hline
\multirow{4}{*}{\textbf{GAD-7}} & LSTM-Concatenation                                                       & -2.68                                                                     \\ \cline{2-3} 
                                & \begin{tabular}[c]{@{}c@{}}BiLSTM-Static Attention\end{tabular}        & -0.84                                                                     \\ \cline{2-3} 
                                & \begin{tabular}[c]{@{}c@{}}LSTM-Tensor Fusion\end{tabular}             & 0.16                                                                      \\ \cline{2-3} 
                                & \textbf{\begin{tabular}[c]{@{}c@{}}CNN-Dynamic Attention\end{tabular}} & \textbf{2.27}                                                             \\ \hline
\multirow{4}{*}{\textbf{SHAPS}} & LSTM-Concatenation                                                       & -1.77                                                                     \\ \cline{2-3} 
                                & \begin{tabular}[c]{@{}c@{}}BiLSTM-Static Attention\end{tabular}        & 0.17                                                                      \\ \cline{2-3} 
                                & \begin{tabular}[c]{@{}c@{}}LSTM-Tensor Fusion\end{tabular}             & 0                                                                         \\ \cline{2-3} 
                                & \textbf{\begin{tabular}[c]{@{}c@{}}CNN-Dynamic Attention\end{tabular}} & \textbf{1.88}                                                             \\ \hline
\end{tabular}
\vskip -0.1in
\end{table}

\subsection{Feature Importance}

We computed SHAP score for the input features, ranked those by scores, and then used the top $5, 10, 25, 50$ and $ 75\% $ features for secondary  modeling (Table \ref{table3}). Model performance remained fairly stable even at a low number of features. For example, for each scale, models built using only 25\% of the original set of features achieved test F1 scores within 3\% of the models built using all features. Specifically for PHQ-9, the difference between the highest F1 (with 75\% of features) and the lowest (with 5\% of features) was only 1\%. These results suggest that a low number of features contributed heavily to the model performance and shows that the SHAP method can effectively identify them. 

Despite differences in the number of features for each modality, the top 10 SHAP-identified most important features for each scale had representation from each modality (Table \ref{table4}). 
Several features co-occured across multiple disorders including text features like Polarity, Subjectivity, Valence and Number of characters and video features including Facial Action Units concerning lip movement. Additionally, there were related audio features that co-occurred across multiple disorders viz. spectral features (Contrast Spectrogram, MFCC and Chroma Energy Normalized Spectrogram) and phonation features (Shimmer and Logarithimic Energy). However, several features were specific to mood disorders e.g. affective text features (Arousal and Dominance) for PHQ-9 and Noun tags for SHAPS (see Supplement for additional results on dynamic SHAP scores). These results potentially highlight some of the shared and divergent themes of these different mood disorders.

\begin{table}[t] 
\centering
\caption{Multimodal classification using reduced number of features selected by SHAP scores: Median Test F1 Score}
\label{table3}
\vskip 0.15in
\begin{tabular}{|c|c|c|c|}
\hline
\textbf{\begin{tabular}[c]{@{}c@{}}\% of total\\ features used\end{tabular}} & \textbf{PHQ-9} & \textbf{GAD-7} & \textbf{SHAPS} \\ \hline
5                                                                            & 0.657          & 0.542          & 0.563          \\ \hline
10                                                                           & 0.654          & 0.588          & 0.580          \\ \hline
25                                                                           & 0.656          & 0.620          & 0.575          \\ \hline
50                                                                           & 0.661          & \textbf{0.638} & 0.595          \\ \hline
75                                                                           & \textbf{0.667} & 0.628          & \textbf{0.599} \\ \hline
100                                                                          & 0.664          & 0.623          & 0.590          \\ \hline
\end{tabular}
\vskip -0.1in
\end{table}

\begin{table}[t] 
\centering
\caption{Top 10 SHAP-Ranked Features for each symptom scale}
\label{table4}
\vskip 0.15in
\small
\begin{tabular}{ lll}
\hline

 \multicolumn{1}{|l|}{\textbf{PHQ-9} }                                           & \multicolumn{1}{l|}{\textbf{GAD-7}}                                                                                                                        & \multicolumn{1}{l|}{\textbf{SHAPS}}                                             \\ \hline
 \multicolumn{1}{|l|}{\cellcolor[HTML]{FFCE93}\begin{tabular}[c]{@{}l@{}} AU16: \\
 Lower Lip \\ Depressor \end{tabular}}     & \multicolumn{1}{l|}{\cellcolor[HTML]{81D980}Polarity}                                                                                            & \multicolumn{1}{l|}{\cellcolor[HTML]{81D980}Polarity}                  \\ \hline
 \multicolumn{1}{|l|}{\cellcolor[HTML]{81D980}Valence}                & \multicolumn{1}{l|}{\cellcolor[HTML]{76C8DF}\begin{tabular}[c]{@{}l@{}}Short-Time \\ Fourier Transform \\ Energy\\ Spectrogram 9\end{tabular}} & \multicolumn{1}{l|}{\cellcolor[HTML]{81D980}Subjectivity}              \\ \hline
 \multicolumn{1}{|l|}{\cellcolor[HTML]{81D980}Dominance}              & \multicolumn{1}{l|}{\cellcolor[HTML]{76C8DF}\begin{tabular}[c]{@{}l@{}}Chroma Energy\\ Normalized \\Spectrogram 11 \end{tabular}}                                                                         & \multicolumn{1}{l|}{\cellcolor[HTML]{FFCE93}\begin{tabular}[c]{@{}l@{}} AU10: \\ Upper Lip \\ Raiser \end{tabular}}           \\ \hline
 \multicolumn{1}{|l|}{\cellcolor[HTML]{81D980}Subjectivity}           & \multicolumn{1}{l|}{\cellcolor[HTML]{81D980}Valence}                                                                                             & \multicolumn{1}{l|}{\cellcolor[HTML]{76C8DF}MFCC 1}                    \\ \hline
 \multicolumn{1}{|l|}{\cellcolor[HTML]{81D980}Polarity}               & \multicolumn{1}{l|}{\cellcolor[HTML]{76C8DF}Shimmer}                                                                                             & \multicolumn{1}{l|}{\cellcolor[HTML]{76C8DF}Spectral Spread}           \\ \hline
 \multicolumn{1}{|l|}{\cellcolor[HTML]{81D980}Arousal}                & \multicolumn{1}{l|}{\cellcolor[HTML]{76C8DF}MFCC 8}                                                                                              & \multicolumn{1}{l|}{\cellcolor[HTML]{81D980}\begin{tabular}[c]{@{}l@{}} Proper \\
 Noun Tag \end{tabular}}             \\ \hline
 \multicolumn{1}{|l|}{\cellcolor[HTML]{76C8DF}MFCC 10}                & \multicolumn{1}{l|}{\cellcolor[HTML]{76C8DF}\begin{tabular}[c]{@{}l@{}}Contrast \\ Spectrogram 3 \end{tabular}}                                           & \multicolumn{1}{l|}{\cellcolor[HTML]{76C8DF}\begin{tabular}[c]{@{}l@{}}Logarithmic \\ Energy \end{tabular}}                    \\ \hline
 \multicolumn{1}{|l|}{\cellcolor[HTML]{81D980}\# of Characters}       & \multicolumn{1}{l|}{\cellcolor[HTML]{76C8DF}MFCC 1}                                                                                              & \multicolumn{1}{l|}{\cellcolor[HTML]{81D980}\begin{tabular}[c]{@{}l@{}}Singular\\ Noun Tag \end{tabular}}                   \\ \hline
             \multicolumn{1}{|l|}{\cellcolor[HTML]{81D980}\begin{tabular}[c]{@{}l@{}} Coordinating \\
 Conjunction\\ Tag\end{tabular}}                   & \multicolumn{1}{l|}{\cellcolor[HTML]{FFCE93}\begin{tabular}[c]{@{}l@{}} AU16: \\
 Lower Lip \\ Depressor \end{tabular}}                                                                     & \multicolumn{1}{l|}{\cellcolor[HTML]{76C8DF}\begin{tabular}[c]{@{}l@{}}Contrast \\ Spectrogram 6 \end{tabular}}  \\ \hline
 \multicolumn{1}{|l|}{\cellcolor[HTML]{76C8DF}\begin{tabular}[c]{@{}l@{}}Contrast \\ Spectrogram 5 \end{tabular}}  & \multicolumn{1}{l|}{\cellcolor[HTML]{81D980}\# of Characters}                                                                                    & \multicolumn{1}{l|}{\cellcolor[HTML]{76C8DF}\begin{tabular}[c]{@{}l@{}}Chroma Energy\\ Normalized \\Spectrogram 4 \end{tabular}} \\ \hline
\multicolumn{1}{l}{}                       &                                                                     &                                                                                                                                                                                                               \\ \hline
 \multicolumn{1}{|c|}{\cellcolor[HTML]{76C8DF}\textbf{Audio}}         & \multicolumn{1}{c|}{\cellcolor[HTML]{FFCE93}\textbf{Video}}                                                                                      & \multicolumn{1}{c|}{\cellcolor[HTML]{81D980}\textbf{Text}}             \\ \hline
\end{tabular}
\vskip -0.1in
\end{table}

\section{Conclusions}

We present a novel, interpretable, multimodal classification method to identify the symptoms of depression, anxiety and anhedonia using audio, video and text data collected from a smartphone app. Key advances of our framework involved using dynamic (i.e. temporal) embeddings for individual modalities and then fusing these using a transformer encoder. This strategy outperformed prior state of the art methods that rely on static (i.e. non-temporal) embeddings in overall F1 scores as well as capturing supplementary multimodal information. Lastly, we used a SHAP-based method to recover the small number of multimodal features that had outsized contributions to the models for clinical validation and digital marker discovery. These analyses were performed on data collected from smartphones outside of clinical/laboratory settings, and further required quality control procedures (which could be incorporated into the design and implementation of future remote studies).

There are several factors to consider for the broader application of our methodology. We used self-reported clinical questionnaires to define the symptom labels, build the models and identify important features. The generalization of these models could be confirmed in independent cohort(s), and the identified features could be validated using a V3 framework \cite{goldsack2020}. Additionally, formally integrating age and gender in our models could improve generalizability.  Nevertheless, we believe that our work will foster other scalable tools focusing on interpretable multimodal understanding of mental health disorders and their symptomatology.

\section*{Acknowledgement}

This work is funded by BlackThorn Therapeutics.

\bibliography{main_paper.bib}
\bibliographystyle{icml2021}

\beginsupplement

\section*{Supplement}

\section{Feature Extraction and Quality Control}

\subsection{Feature Extraction}

\subsubsection{Audio Features}
For each audio file, 123 audio features were extracted from the voiced segments at a resolution of 0.1 seconds, including prosodic (Pause rate, speaking rate etc.), glottal (Normalised Amplitude Quotient, Quasi-Open-Quotient etc.), spectral (Mel-frequency cepstral coefficients, Spectral Centroid, Spectral flux, Mel-frequency cepstral coefficient spectrograms etc.) and chroma (Chroma Spectogram) features \cite{cummins2015}.

\subsubsection{Video Features}
For each video file, 22 Facial Action Unit features were extracted. These were derived from 3D facial landmarks which were computed at a resolution of 0.1 seconds. This was in contrast to prior literature where 2D facial landmarks have been primarily used. We found in our experiments that 3D facial landmarks were much more robust to noise than 2D facial landmarks, thus making these ideal for remote data collection. 

\subsubsection{Text Features}

For each file, 52 text features were extracted including affect based features viz. arousal, valence and dominance rating for each word using Warriner Affective Ratings \cite{Warriner13}, polarity for each word using TextBlob \cite{blob}, contextual features such as word embeddings using doc2vec \cite{le14} etc.

\subsection{Quality Control} 

We built several quality control flags for screening participant sessions using collected metadata, scales and extracted features. These included flags on

(1) Video frame capture failures (poor lighting conditions)

(2) Missing transcriptions (excessive background noise or multiple persons speaking)

(3) Illegible speech

(4) Inconsistent responses between similar questions of clinical scales
    
These flags were used to remove violating participant sessions. 

\section{Model} \label{model}

\subsection{Model Architecture}
The model architecture comprised of two components:

\subsubsection{Unimodal Encoders}

\begin{figure}[ht] 
\centering
    \includegraphics[width=1.0\columnwidth]{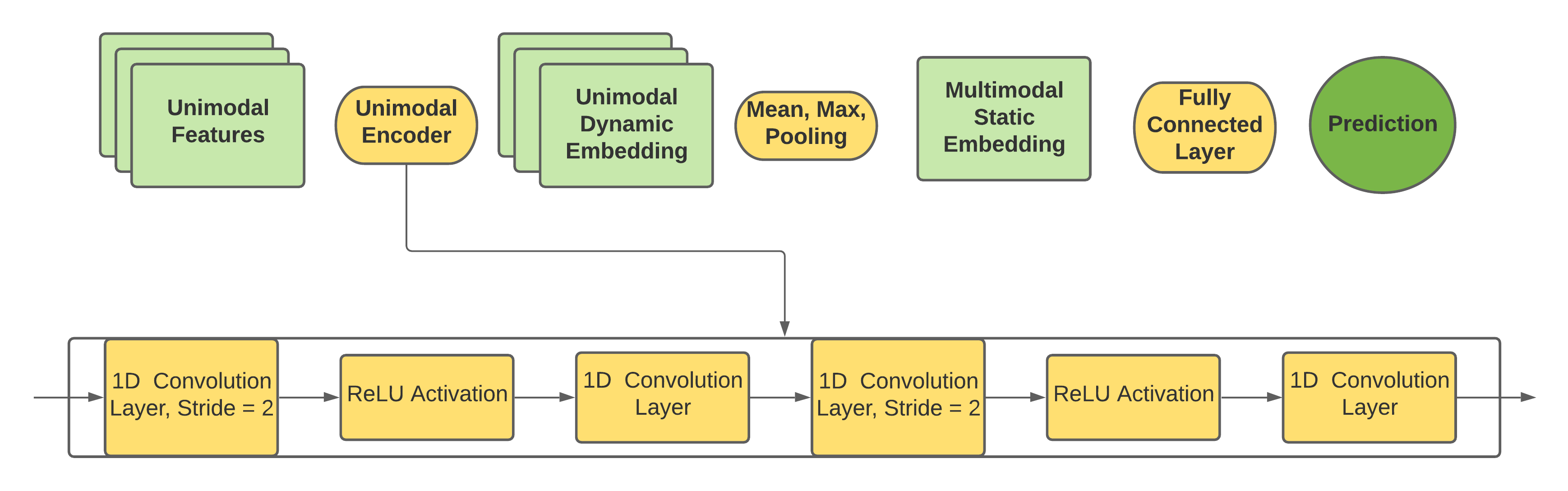}
    \caption{Schematic of Unimodal Encoder.}
    \label{figuni}
\end{figure}

The unimodal encoder (schematic in Figure \ref{figuni}) comprised of stacked  convolution blocks followed by fully-connected layers. Each convolution block was composed of a 1D convolution layer (that performed strided 1D convolution) followed by ReLU activation function. The unimodal encoders were each individually pre-trained on the symptom labels before use in the full, multimodal system. For pre-training, the output of the final block was flattened through max and mean pooling, these two vectors were concatenated and input to a fully-connected layer followed by sigmoid activation. After pre-training, the pooling and the fully connected layers were discarded and the output dynamic embedding was passed on to the fusion layer.

\subsubsection{Transformer based Multimodal Fusion}

The dynamic unimodal embeddings of the previous step were concatenated and passed on to a Transformer Encoder \cite{vaswani17} , comprising of stacked transformer encoder blocks. Each transformer encoder block was composed of a Multi-Head Self Attention and a feedforward network. The output of the Transformer Encoder was flattened through max and mean pooling, these two vectors were concatenated and passed on to a fully-connected layer for classification.

\subsection{Model Training and Evaluation}
For each of the three binary classification problems, random search was used to do hyperparameter tuning using the validation set. 
For the unimodal encoder pre-training, the following hyperparameters were tuned: the number of CNN layers, the number of hidden nodes in each layer, the kernel size, the number and size of fully connected layers. Dropout was applied to all non-linear layers with a 0.2 probability of being zeroed. The Adam optimizer with L2 weight decay and binary cross-entropy loss was used. The model was trained for 30 epochs. After pre-training, the pooling and the fully connected layers was discarded and an output dynamic embedding was passed on to the fusion layer.
For the multimodal training, the following hyperparameters were tuned: the number of transformer encoder layers, the number of hidden nodes and the number of multihead attention heads in the transformer layers, the number and size of fully connected layers. Dropout was applied to the fully connected layers with a 0.4 probability of being zeroed. The Adam optimizer with L2 weight decay and binary cross-entropy loss was used. The model was trained for 30 epochs.
The entire model was implemented in PyTorch.

For training and evaluating the model, we used the F1 score as the metric. This was particularly important given the unbalanced nature of the class labels.

\subsection{SHAP-based Feature Importance}


We employed the SHAP Gradient Explainer algorithm which is an extension of the Integrated Gradients algorithm \cite{Sundararajan2017}.
This allowed us to calculate SHAP scores for each of our input variables for each subject and time point. 
We summarized these values over the subjects and time points, resulting in a global importance value for each feature.

\section{Experiments and Results}

\subsection{Multimodal Classification of Symptom Severity}

We trained models for each modality (audio, video and text) and compared the performance of the best performing unimodal model with the trimodal model for each approach and scale. The complete results are shown in Table \ref{table5}. To train models with only one modality, after learning the dynamic unimodal embedding using the 1D-CNN based encoder, the output was flattened through max and mean pooling, these two vectors were concatenated and input to a fully-connected layer (with dropout) followed by softmax activation to generate the output probability and trained using backpropagation.

\subsection{Feature Importance}

\begin{table*}[] 
\centering
\caption{Comparison between median test F1 scores for unimodal vs trimodal models}
\label{table5}
\vskip 0.15in
\begin{tabular}{|c|c|c|c|c|c|c|}
\hline
\textbf{Scale}                  & \textbf{Approach}              & \textbf{\begin{tabular}[c]{@{}c@{}}Audio\\ Performance\end{tabular}} & \textbf{\begin{tabular}[c]{@{}c@{}}Video \\ Performance\end{tabular}} & \textbf{\begin{tabular}[c]{@{}c@{}}Text\\ Performance\end{tabular}} & \textbf{\begin{tabular}[c]{@{}c@{}}Trimodal\\ Performance\end{tabular}} & \textbf{\begin{tabular}[c]{@{}c@{}}Percentage \\
Difference\\ between\\ Multimodal \&\\ Best Unimodal \\
Model\end{tabular}} \\ \hline
\multirow{4}{*}{\textbf{PHQ-9}} & LSTM-Concatenation             & 0.566                                                                & 0.568                                                                 & 0.633                                                               & 0.628                                                                   & -0.79                                                                                                                  \\ \cline{2-7} 
                                & BiLSTM-Static Attention        & 0.565                                                                & 0.572                                                                 & 0.629                                                               & 0.629                                                                   & 0                                                                                                                      \\ \cline{2-7} 
                                & LSTM-Tensor Fusion             & 0.548                                                                & 0.559                                                                 & 0.632                                                               & 0.633                                                                   & 0.16                                                                                                                   \\ \cline{2-7} 
                                & \textbf{CNN-Dynamic Attention} & 0.579                                                                & 0.587                                                                 & 0.644                                                               & 0.664                                                                   & \textbf{3.05}                                                                                                          \\ \hline
\multirow{4}{*}{\textbf{GAD-7}} & LSTM-Concatenation             & 0.544                                                                & 0.560                                                                 & 0.604                                                               & 0.588                                                                   & -2.68                                                                                                                  \\ \cline{2-7} 
                                & BiLSTM-Static Attention        & 0.546                                                                & 0.572                                                                 & 0.600                                                               & 0.595                                                                   & -0.84                                                                                                                  \\ \cline{2-7} 
                                & LSTM-Tensor Fusion             & 0.552                                                                & 0.564                                                                 & 0.600                                                               & 0.601                                                                   & 0.16                                                                                                                   \\ \cline{2-7} 
                                & \textbf{CNN-Dynamic Attention} & 0.547                                                                & 0.570                                                                 & 0.609                                                               & 0.623                                                                   & \textbf{2.27}                                                                                                          \\ \hline
\multirow{4}{*}{\textbf{SHAPS}} & LSTM-Concatenation             & 0.536                                                                & 0.518                                                                 & 0.567                                                               & 0.557                                                                   & -1.77                                                                                                                  \\ \cline{2-7} 
                                & BiLSTM-Static Attention        & 0.544                                                                & 0.519                                                                 & 0.564                                                               & 0.565                                                                   & 0.17                                                                                                                   \\ \cline{2-7} 
                                & LSTM-Tensor Fusion             & 0.541                                                                & 0.517                                                                 & 0.567                                                               & 0.567                                                                   & 0                                                                                                                      \\ \cline{2-7} 
                                & \textbf{CNN-Dynamic Attention} & 0.571                                                                & 0.517                                                                 & 0.579                                                               & 0.590                                                                   & \textbf{1.88}                                                                                                          \\ \hline
\end{tabular}

\vskip -0.1in
\end{table*}

\begin{figure*} 
\centering
    \includegraphics[width=2\columnwidth]{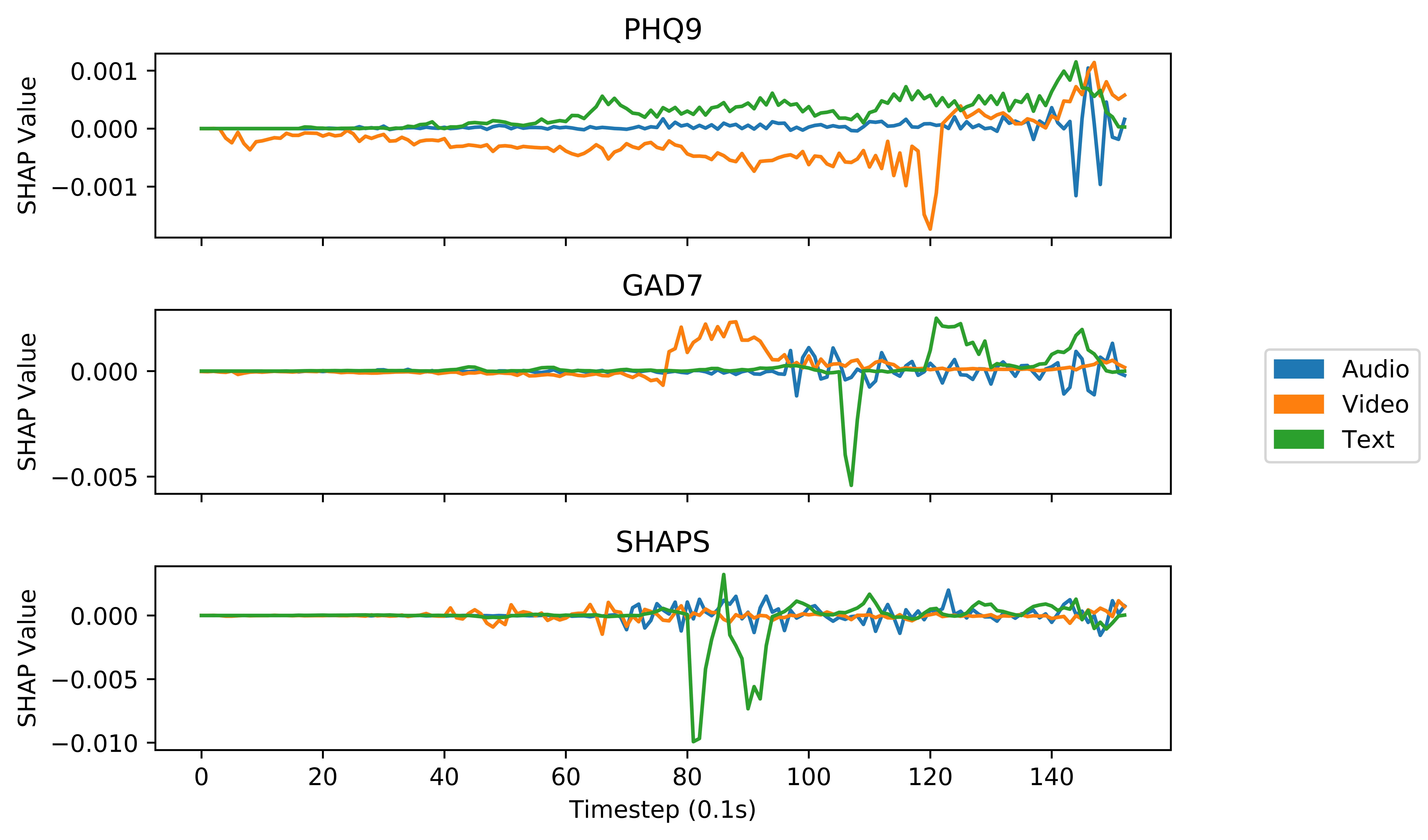}
    \centering\caption{SHAP Value over time for the top ranked feature of each modality}
    \label{figSHAP}
\end{figure*}

The SHAP-based interpretability method enabled us to compute the
SHAP values for all the input features (which are time series features) over each point of time. Thus we could get an additional level of information for the feature importance. For example, clinicians could use these type of data to isolate the impacts of specific words, facial or vocal expressions, or to find particularly salient portions of the video in larger clips. As a demonstration, we show the time series SHAP values of one participant for the top audio, video and text features for each of the scales in Figure \ref{figSHAP}.

\end{document}